\documentclass{article}

\usepackage[a4paper,
            bindingoffset=0.2in,
            left=1in,
            right=1in,
            top=1in,
            bottom=1in,
            footskip=.25in]{geometry}
            
\usepackage[utf8]{inputenc} 
\usepackage[T1]{fontenc}    
\usepackage{hyperref}       
\usepackage{url}            
\usepackage{booktabs}       
\usepackage{amsfonts}       
\usepackage{nicefrac}       
\usepackage{microtype}      
\usepackage{xcolor}         
\usepackage{amsmath}
\usepackage{graphicx}
\usepackage{physics}
\usepackage{amsthm}
\usepackage{wrapfig,booktabs}

\usepackage{titlesec}

\newcommand{\be}{\begin{equation}}
\newcommand{\ee}{\end{equation}}
\newcommand\myeq{\mathrel{\stackrel{\makebox[0pt]
{\mbox{\normalfont\tiny def}}}{=}}}

\title{High-performance deep spiking neural networks\\ with 0.3 spikes per neuron}
\author{
  \textbf{Ana Stanojevic}$^{1, 2}$ \hspace{1cm}  \textbf{Stanis{\l}aw Wo{\'z}niak}$^{1, \footnote{Correspondence to: Stanis{\l}aw Wo{\'z}niak.
E-mail address: stw@zurich.ibm.com}} $  \hspace{1cm} \textbf{Guillaume Bellec}$^{2}$  \\  \textbf{Giovanni Cherubini}$^{1}$ \hspace{1cm} \textbf{Angeliki Pantazi}$^{1}$  \hspace{1cm} \textbf{Wulfram Gerstner}$^{2}$   \\
   ${^{1}}$ {IBM Research Europe – Zurich, R{\"u}schlikon, Switzerland} \\
  ${^{2}}$ {\'E}cole polytechnique f{\'e}d{\'e}rale de Lausanne, Lausanne EPFL, Switzerland \\
}
\date{}

\begin{document}
\maketitle

\begin{abstract}
Communication by rare, binary spikes is a key factor for the energy efficiency of biological brains. However, it is harder to train biologically-inspired spiking neural networks (SNNs) than artificial neural networks (ANNs). This is puzzling given that theoretical results provide exact mapping algorithms from ANNs to SNNs with  time-to-first-spike (TTFS) coding.  In this paper we analyze in theory and simulation the learning dynamics of TTFS-networks and identify a specific instance of the vanishing-or-exploding gradient problem. While two choices of SNN mappings solve this problem at initialization, only the one with a constant slope of the neuron membrane potential at threshold guarantees the equivalence of the training trajectory between SNNs and ANNs with rectified linear units. We demonstrate that training deep SNN models achieves the exact same performance as that of ANNs, surpassing previous SNNs on image classification datasets such as MNIST/Fashion-MNIST, CIFAR10/CIFAR100 and PLACES365. Our SNN accomplishes high-performance classification with less than 0.3 spikes per neuron, lending itself for an energy-efficient  implementation. We show that fine-tuning SNNs with our robust gradient descent algorithm enables their optimization
for hardware implementations with low latency and resilience to noise and quantization.
\end{abstract}
\section{Introduction}

Similar to the brain,  neurons in  spiking neural networks (SNNs) communicate via short pulses called spikes that arrive in continuous time -- in striking contrast to   artificial neural networks (ANNs) where neurons communicate by the exchange of real-valued signals in discrete time. While ANNs are the basis of modern artificial intelligence with impressive achievements \cite{gpt3, perceiver, yu2021}, their high performance on various tasks comes at  the expense of high energy consumption \cite{energy_server, Patterson22, sustainable_ai}. In general, high energy consumption is a challenge in terms of sustainability and deployment in low-power edge devices \cite{dl_edge_merge, google_nn, ai_on_edge}. SNNs may offer a potential solution due to their sparse binary communication scheme  that reduces the resource usage in the network \cite{neuromorphic_in_memory1, neuromorphic_in_memory2, Goltz21,gallego2020event, Davies21, Diehl16}; however, it has been so far impossible to train  deep SNNs that perform at the exact same level as ANNs.

Multiple methods have been proposed to train the parameters of SNNs. 
Traditionally, they were trained with plasticity rules observed in biology \cite{masquelier2007unsupervised,kheradpisheh2018stdp}, but it appears more efficient to rely on gradient-descent optimization ('Backprop') as done in deep learning \cite{Goodfellow16}.
One of the most successful training paradigms for SNNs views each spiking neuron as a discrete-time recurrent unit with binary activation and uses a pseudo-derivative or surrogate gradient on the backward pass while keeping the strict threshold function in the forward pass \cite{Neftci19,bellec2018long,Zenke18,learning_gd_snu,learning_gd_approx}. Other approaches \cite{schmitt2017neuromorphic, learning_gd_prob, stanojevic2023} either translate the ANN activations into SNN spike counts to train the SNN with the ANN gradients, or use temporal coding with a large number of spikes. Both jeopardize the energy efficiency, because  the number of spikes is directly related to energy consumption, in digital~\cite{Davies21,han2015deep} as well as in analog hardware or mixed analog-digital neuromorphic systems \cite{Goltz21}. 

In contrast to spike-count measures or rate coding in neuroscience \cite{hubel1959} and real-valued signals in ANNs \cite{Goodfellow16}, it was found in sensory brain areas that neurons also encode information in the exact timing of the {\em first} spike, i.e.  more salient information leads to earlier spikes \cite{temporal_retina,temporal_tactile,Carr93}
 which in turn leads to a fast response to stimuli \cite{ Thorpe89,Thorpe96,Thorpe01}. Specifically, we focus on the {\em time-to-first-spike} (TTFS) coding scheme \cite{Maass97,Gerstner98b,Maass98}, in which each neuron fires at most a single spike. With less than one spike per neuron, SNNs with TTFS coding (TTFS-networks), are  an excellent choice for energy-efficient inference. 

 For future implementations of high-performance deep TTFS-networks, a critical piece of the puzzle is, however, still missing. 
Energy-optimized hardware always comes with constraints such as weight quantization or sparsity in digital hardware \cite{Davies21,han2015deep,liu2018memory,courbariaux2015binaryconnect}, and parameter mismatch or noise in analog hardware \cite{schmitt2017neuromorphic,wunderlich2019demonstrating}. 
The known solutions for addressing these constraints are either to train from scratch a custom hardware-specific model or to initialize with a converted model and fine-tune to fit the hardware constraints \cite{schmitt2017neuromorphic}. In both cases we need parameter optimization algorithms.
However, in TTFS-networks none of the known gradient-descent learning algorithms \cite{Goltz21,Bohte2002,wunderlich2021,zhang21,mostafa,comsa} is robust enough to generalize to deep neural networks making these standard training pipelines impracticable for spiking neuromorphic hardware.

Training TTFS-networks with gradient-descent has a long history \cite{Bohte2002}.  Using the Spike Response Model \cite{Gerstner98b} it is possible to calculate backpropagation gradients with respect to spike timing and parameters \cite{Bohte2002}. While the original paper states that the learning rule contains an approximation, it turns out to be the exact gradient when the number of spikes is fixed, i.e. the presence or absence of a spike per neuron remains unchanged \cite{Goltz21,wunderlich2021,zhang21,mostafa,comsa,stanojevic2020}.
However, unless ad-hoc gradient approximations are introduced 
\cite{park2021training, zhou2021temporal},
none of these theoretically sound studies could train a spiking network with more than six layers to high performance.
An alternative that avoids the training altogether is to convert directly an ANN into an SNN, using either  rate coding  \cite{rate_snn_bodo, Hu2021} or temporal coding in  SNNs  \cite{Maass97,Maass95,Stockl21, Bu22, Rueckauer18, stanojevic2022}. While most of the conversions relied on approximate mapping algorithms, it was recently shown that an approximation-free conversion from an ANN with rectified linear units (ReLU-network) to a TTFS-network is possible \cite{stanojevic2022}. In spite of the existence of a mapping between ANNs and SNNs \cite{Maass97,stanojevic2022}, training or fine-tuning deep SNNs with gradient descent in the TTFS setting has remained challenging, suggesting that unknown difficulties arise during spike-time optimization.

Here we theoretically analyze why training deep TTFS-networks has encountered difficulties in closing the gap in performance compared to ANNs, and we provide a solution that closes this gap.
Our approach relies on the combination of exact backpropagation updates~\cite{Bohte2002, wunderlich2021,zhang21,mostafa, stanojevic2020}  with an exact revertible mapping between ReLU-networks and TTFS-networks inspired by \cite{stanojevic2022}.  Together, these two ingredients enable the following contributions. Firstly, we identify analytically that SNN training is typically unstable due to a severe vanishing-or-exploding gradient problem \cite{bengio_vanishing, Hochreiter01, Goodfellow16} which arises when naively using ANN parameter initialization in TTFS-networks. 
Secondly, we explain why even with corrected parameter initialization and exact gradient updates the performance of a trained TTFS-network is typically worse than that of the corresponding ReLU network. 
We identify a specific TTFS-network parameterization ('identity mapping') that ensures an even stricter condition, i.e., gradient descent in the SNN follows the same learning trajectories as in the equivalent ReLU-network. 
Thirdly, implementing these theoretical considerations enables TTFS-networks to be trained to the exact same accuracy as deep ReLU-networks on various standard image classification datasets, including MNIST, Fashion-MNIST, CIFAR10, CIFAR100, and PLACES365, and fine-tuned to operate with less than 0.3 spikes per neuron. 
Our results surpass the performance of all previous SNNs, including those that relied on approximations of gradients or  mappings
\cite{Goltz21,mostafa,  comsa, zhang21, stanojevic2020, park2021training, zhou2021temporal,Stockl21,Bu22, Rueckauer18}. 
Our approach paves the way to convert high-performance pre-trained ANNs to TTFS-networks and fine-tune them to the specific hardware characteristics while optimizing for low latency or minimizing energy by reducing the number of spikes per neuron.

\section{Results}
\subsection{The SNN architecture with TTFS coding} 
\label{sec:second}

\begin{figure*}[!t]
\centering
\includegraphics{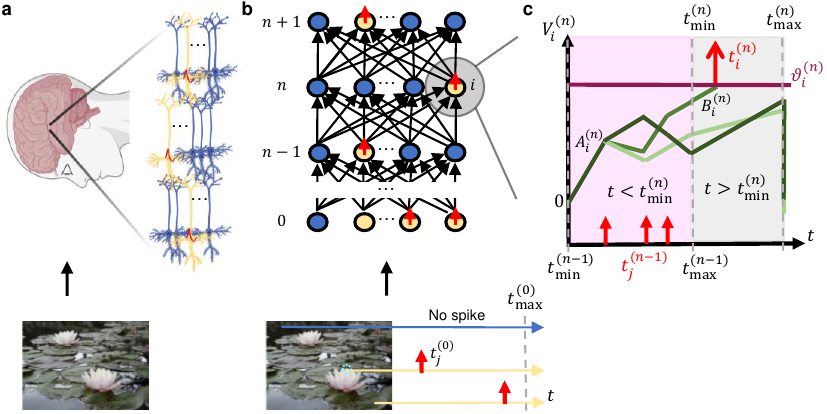}
\caption{{\bf Network of TTFS neurons.} {\bf a.} An image (bottom) is processed in the brain through rapid transmission of spikes between neurons \cite{Thorpe96,Richmond90}. Yellow color denotes active neurons, and red color represents the generated spikes (schematic). {\bf b.} A feed-forward TTFS-network architecture receives spikes, where earlier spiking times encode more salient information in the image \cite{temporal_retina}, e.g., receptive fields around a lotus flower (dashed circle, bottom). {\bf c.} The potential $V_i^{(n)}$ for different neurons $i$ varies as a function of time. The initial slope is $A_i^{(n)}$ (here the same value for all neurons). In the first regime $t<t_{\rm min}^{{(n)}}$ (pink), the slope changes as the input spikes arrive, and in the second regime $t>t_{\rm min}^{{(n)}}$ (gray), the slope is fixed at $B_i^{(n)}$. The neuron spikes if its potential reaches the spiking threshold $\vartheta_i^{(n)}$ before $t_{\text{max}}^{(n)}$. 
} 
\label{fig:Fig1}
\end{figure*}

Inspired by the fast processing \cite{Thorpe89,Thorpe96} in the brain (Fig. \ref{fig:Fig1}a), we study deep SNNs  consisting of neurons which are arranged in $N$ hidden layers where the spikes of neurons in layer $n$ are sent to  neurons in layer $n+1$ (Fig. \ref{fig:Fig1}b).  The layers are either fully-connected (i.e. each neuron receives input from all neurons in the previous layer) or convolutional (i.e. connections are limited to be local and share weights). 
In the following equations, an upper index refers to the layer number  while column vectors, denoted in boldface, refer to all neurons in a given layer.

 The $D$ real-valued inputs, such as pixel intensities, are first scaled to the interval $[0, 1]$, resulting in the vector $\mathbf{x}^{(0)}=(x_1^{(0)}, x_2^{(0)}, \dots, x_D^{(0)})^T$ where $x_j^{(0)}\in [0, 1]$, and then encoded into spiking times of the SNN input layer using TTFS coding (Fig. \ref{fig:Fig1}b). A high input pixel intensity leads to an early spike at time $t_j^{(0)} = t_{\rm max}^{{(0)}} - \tau_c x_j^{(0)}$, and specifically for $x_j^{(0)}=0$ no spike will be emitted. 
 The conversion parameter $\tau_c$ translates unit-free inputs into time units. 
The potential $V_i^{(n)} (t)$ of a neuron $i$ in hidden layer $n\ge 1$ is initialized at zero and described by integrate-and-fire dynamics with linear post-synaptic potential (Fig. \ref{fig:Fig1}c), which we view as a linearization of a classical double-exponential filter~\cite{Goltz21,comsa}, see \ref{ap:generalization}. 
Given the spike times $t_j^{(n-1)}$ of neurons $j$ in the previous layer, the potential $V_i^{(n)}$ follows the dynamics:
\begin{equation}\label{eq:potential}
\tau_c \dv{V_i^{(n)}}{t} = \left\{
\begin{split}
\; A_i^{(n)}&+
\sum_j W_{ij}^{(n)}H(t-t_j^{(n-1)}) &  \qquad \text{\rm for \quad}  t < t_{\text{min}}^{(n)} \\ 
\; B_i^{(n)}&  
\nonumber
&  \qquad \text{\rm for \quad} t_{\text{min}}^{(n)} \le t \le t_{\text{max}}^{(n)} 
\end{split}
\qquad \right. (1)
\end{equation}
where the $t_{\rm min}^{(n)}$ and $t_{\rm max}^{(n)}$ are temporal bounds separating two regimes of membrane potential behaviour (Fig. \ref{fig:Fig1}c). 
 Here $B_i^{(n)}>0$  and $A_i^{(n)}$ are scalar parameters, $W_{ij}^{(n)}$ is the synapse strength from neuron $j$ to neuron $i$ , and $H$ denotes the Heaviside function which takes a value of $1$ for positive arguments and is $0$ otherwise. 
In the first regime, for $t<t_{\text{min}}^{(n)}$, the slope of the voltage trajectory starts with a value $A_i^{(n)}$ and  increases or decreases after each spike arrival, depending on the sign of $W_{ij}^{(n)}$, whereas at the time $t=t_{\text{min}}^{(n)}$ the potential enters the second regime, switching the slope to a fixed positive value $B_i^{(n)}$. When the potential $V_i^{(n)}$ reaches the threshold $\vartheta_i^{(n)}$,  neuron $i$ generates a spike at time $t_i^{(n)}$ and sends it to the next layer. The threshold $\vartheta_i^{(n)}$ is defined as $\vartheta_i^{(n)} \myeq \Tilde{\vartheta}_i^{(n)} + D_i^{(n)}$ where $D_i^{(n)}$ is a trainable parameter initialized at 0 and $\Tilde{\vartheta}_i^{(n)}$ is a fixed base threshold, 
defined to be large enough to prevent firing before $t_{\rm min}^{(n)}$, see \ref{ap:ttfs_details}. 
Importantly, we also define $t_{\rm max}^{(n)}$, such that for $t\ge t_{\rm max}^{(n)}$ emission of a spike is impossible, e.g., implemented by a de-charging current for  $t\ge t_{\rm max}^{(n)}$ which resets the membrane potential to zero (Fig. \ref{fig:Fig1}c). 
Once a neuron spikes we assume a long refractory period to ensure that every neuron spikes at most once. 
The construction of $t_{\rm min}^{(n)}$ and $t_{\rm max}^{(n)}$ is recursive with $t_{\rm min}^{(n)} \myeq t_{\rm max}^{(n-1)}$. 

\addtocounter{equation}{1}

The model  defined by Eq. \eqref{eq:potential} is rather general and contains several other models as special cases. It is identical to the integrate-and-fire model if we set 
$B_i^{(n)} = A_i^{(n)}+   \sum_j W_{ij}^{(n)}H(t_{\rm min}^{(n)}-t_j^{(n-1)})$; we note that in this case the parameter $B_i^{(n)}$ depends on the sequence of spikes that has arrived from the previous layer. In order to avoid this dependency, an earlier study \cite{stanojevic2022} related
$B_i^{(n)}$ to $A_i^{(n)}$  via an auxiliary parameter
$\alpha_i^{(n)} \myeq  A_i^{(n)} >0$ and 
$B_i^{(n)}\myeq \alpha_i^{(n)} + \sum_j W_{ij}^{(n)}$. 
We use this latter model as a comparison point
with the choice $\alpha_i^{(n)}=1$, and call it the $\alpha1$-model.

It is known that any ReLU network can be mapped to the $\alpha1$-model \cite{stanojevic2022}, but the mapping theory can be extended to our more general model with arbitrary parameters $B_i^{(n)}$.
In the following, we set always
$A_i^{(n)}=0$, but keep $B_i^{(n)}$ arbitrary. 
 We now describe an \textit{exact reverse mapping} which uniquely defines the parameters of an equivalent ReLU-network  (up to the intrinsic scaling symmetry of ReLU units) with the same architecture as that of the SNN. Given the weights $W_{ij}^{(n)}$ and thresholds $\vartheta_{i}^{(n)}$ of the SNN, the weight matrices $w^{(n)}$ and bias vectors $\mathbf{b}^{(n)}$ of the equivalent ReLU network are:
\begin{equation}
\label{eq:general_mapping}
 w_{ij}^{(n)} \myeq \mathcal{M}(W_{ij}^{(n)})\myeq \frac{W_{ij}^{(n)}}{B_i^{(n)}} 
 \text{\hspace{1em} and \hspace{1em}}
 b_{i}^{(n)} \myeq - \frac{\vartheta_i^{(n)}}{B_i^{(n)}} + \frac{t_{\text{max}}^{(n)} - t_{\text{min}}^{(n)}}{\tau_c},
\end{equation}
where $\mathcal{M}$ is a function that maps the weights $W_{ij}^{(n)}$ of the TTFS-network to the weights $w_{ij}^{(n)}$ of the ReLU-network with the parameter $B_i^{(n)}$ defined in Eq. \eqref{eq:potential}. Then, for the vector $\mathbf{x}^{(0)}$ of input activations, the reverse mapping in Eq. \eqref{eq:general_mapping} defines uniquely a ReLU-network with activation vectors $\mathbf{x}^{(n)}$ in layer $n$ such that 
$x_i^{(n)}  =   ({t}_{\rm max}^{(n)} - {t}_i^{(n)})/\tau_c $
for neurons that fire a spike and $x_i^{(n)}=0$ for neurons in the SNN that do not fire a spike
(see Methods for proof). For the non-spiking output layer $N+1$, we allow for  a value $A_i^{(N+1)} \ne 0 $. (Methods). The reverse mapping is then given by  $w_{ij}^{(N+1)} \myeq W_{ij}^{(N+1)}$ and $b_i^{(N+1)} \myeq 
({t}_{\rm max}^{(N)} - {t}_{\rm min}^{(N)}) A_i^{(N+1)}  $, yielding at the readout-time $ {t}_{\rm read}^{(N+1)}= {t}_{\rm max}^{(N)}   $ the same logits and cross-entropy loss $\mathcal{L}$ at the output of the TTFS-network as in the equivalent ReLU-network \cite{stanojevic2022}.

The mapping defined in Eq. \eqref{eq:general_mapping} is a fundamental pillar in the theoretical analysis of the learning dynamics in the next section. Importantly, the parameter $B_i^{(n)}$, which represents the slope of the potential at the moment of threshold crossing (when time is measured in units of $\tau_c$; see Eq. \eqref{eq:potential} and Fig. \ref{fig:Fig1}c), will play a crucial role for learning in SNNs.

\subsection{The vanishing-or-exploding gradient problem in SNNs}

\label{sec:third}

The TTFS-network defined in Eq.~\eqref{eq:potential} is represented in \textit{continuous time} and trained using exact backpropagation, where the derivatives are computed with respect to the spiking times. Previous TTFS-networks, which are trained with exact gradients, primarily utilize shallow architectures with only one hidden layer \cite{Goltz21, comsa, mostafa, stanojevic2020}, or they employ gradient approximations when training deeper networks \cite{park2021training, zhou2021temporal}. The question arises: why does the exact gradient approach not scale well to larger networks? In this section, we demonstrate that deep TTFS-networks generically cause vanishing or exploding gradients, known as the vanishing-gradient problem \cite{Goodfellow16, bengio_vanishing,Hochreiter01}, which we address in the following analysis.

The activity of the SNN at layer $n$ is summarized by the vector of spike timings $\mathbf{t}^{(n)}$ such that the loss with respect to the weights parameters at layer $n$ factorizes as: 
\begin{align}
\label{eq:dL_dt}
\dv{\mathcal{L}}{\mathrm{W}^{(n)}} &= \dv{\mathcal{L}}{ \mathbf{V}^{(N+1)}} \dv{ \mathbf{V}^{(N+1)}}{ \mathbf{t}^{(N)}} \dv{{ \mathbf{t}^{(N)}}}{{ \mathbf{t}^{(N-1)}}} \dots \dv{ \mathbf{t}^{(n+1)}} { \mathbf{t}^{(n)}}  \dv{ \mathbf{t}^{(n)}}{ \mathrm{W}^{(n)}},
\end{align}
where $\mathbf{V}^{(N+1)}$ is a vector containing potentials of neurons in the output layer $N+1$ at time $t_{\text{min}}^{(N+1)}$. 
Formally, for the definition of the firing time vector ${ \mathbf{t}^{(n)}}$, the firing time of non-spiking neurons is set to an arbitrary constant, so that its derivative vanishes. If the product of $\dv{ \mathbf{t}^{(n+1)}} { \mathbf{t}^{(n)}}$ Jacobians is naively defined, the amplitude of this gradient might vanish or explode exponentially fast as the number of layers becomes large.

\begin{figure*}[!t]
\centering
\includegraphics{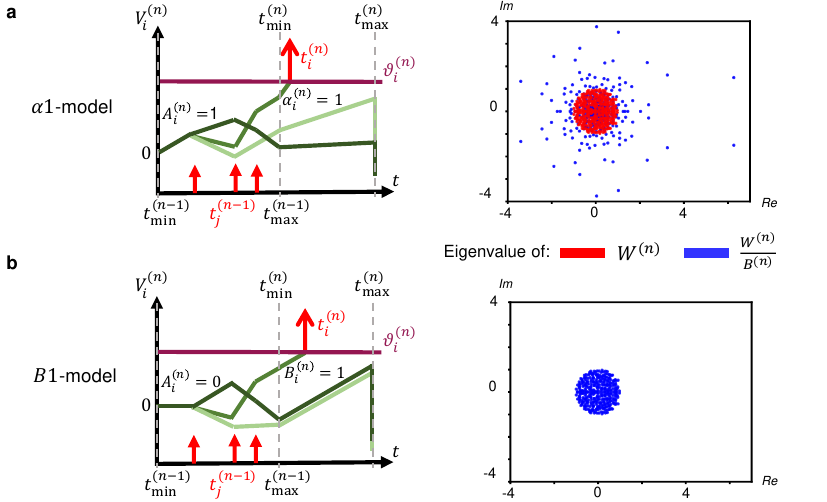}
\caption{{\bf Eigenvalues of the SNN Jacobian for different models under standard deep learning initialization.}  {\bf a.} The $\alpha1$-model. (Left) The initial slope is $A_i^{(n)}=\alpha_i^{(n)}=1$ and in the second regime the slopes of all neurons $i$ are set to neuron-specific slopes derived from $\alpha_i^{(n)}=1$.
(Right) The eigenvalues of the SNN Jacobian spread beyond the unit circle as $B_i^{(n)}\neq1$, i.e. the network will experience exploding gradients at initialization. {\bf b.} The $B1$-model. (Left) The initial slope is $A_i^{(n)}=0$ and in the second regime the slopes of all neurons $i$ are set exactly to $1$. (Right) The eigenvalues of the SNN Jacobian spread inside the unit circle as $B_i^{(n)}=1$, i.e. the network will not experience the exploding gradient at initialization.}
\label{fig:Fig2}
\end{figure*}

To calculate analytically the Jacobian of the SNN, we define a diagonal matrix  $M^{(n)}$ with elements $M_{ij}^{(n)}=\delta_{ij} H(t_{\rm max}^{(n)}-t_i^{(n)})$ that are $1$ if and only if spike $t_i^{(n)}$ occurs before $t_{\rm max}^{(n)}$.
The Jacobian of the network can then be written as ($\cdot$ is the matrix multiplication), see Methods for calculations:
\begin{equation}
\label{eq:dt_dt}
\dv{\mathbf{t}^{(n)}}{\mathbf{t}^{(n-1)}} = M^{(n-1)} \cdot \frac{1}{B^{(n)}} \cdot W^{(n)} = M^{(n-1)} w^{(n)}.
\end{equation} 
where the matrix $B^{(n)}$ is the diagonal matrix with elements $B_i^{(n)}$. From the exact reverse mapping, we know that $M_{ii}^{(n)}$ is 1 if and only if the equivalent ReLU unit $i$ in layer $n$ has a non-zero output. 
In the ANN literature \cite{bengio_vanishing, sussillo2014random,he2015delving}, a method to tackle the problem of vanishing-or-exploding gradients at initialization is to make sure that the largest eigenvalues of  the Jacobian  are close to $1$ in absolute value. Following classical work in the field of ANNs, we assume that $M^{(n-1)}$ has a small impact on the distribution of the eigenvalues of  $\dv{\mathbf{t}^{(n)}}{\mathbf{t}^{(n-1)}}$ \cite{sussillo2014random,he2015delving}.  With this assumption, the focus in a standard ReLU network is  just on  the largest eigenvalue of the weight matrix $w^{(n)}$ \cite{sussillo2014random,he2015delving}. However, in the case of a TTFS-network, a new problem arises due to the fact that the weight matrix $W^{(n)}$ is multiplied by $1/B^{(n)}$; see Eq. \eqref{eq:dt_dt}. Therefore, the eigenvalues of the Jacobian $\dv{ \mathbf{t}^{(n)}} { \mathbf{t}^{(n-1)}}$  are strongly determined by the diagonal matrix $B^{(n)}$ and not only by the weight matrix $W^{(n)}$ of the SNN.
 
In Fig. \ref{fig:Fig2}, we demonstrate numerically that initializing the weight matrix $W^{(n)}$ using standard deep learning recipes can result in vanishing-or-exploding gradients. The weight matrix of an SNN is initialized with $W^{(n)} = \frac{1}{\sqrt{340}} \mathcal{N}(0,1)$ where $340$ is the number of units in each of the eight layers (this is one of the many standard choices in deep learning \cite{Goodfellow16}) so the eigenvalue of $W^{(n)}$ with largest absolute value is close to $1$. We study two models, 
the $\alpha1$-model \cite{stanojevic2022} introduced above
and a model with $B_i^{(n)}=1$ for all neurons and layers that we will call the $B1$-model.
As shown in Fig.~\ref{fig:Fig2}a, the standard deep learning initialization produces multiple eigenvalues with moduli larger than $1$ in a single layer of the $\alpha1$-model, leading in a network of eight layers to an explosion of the gradient norm already at the beginning of the training. 

With this insight, we can now define two different approaches to solve the problem. The first one uses the $\alpha1$-model, but with an  initialization scheme adapted to SNNs. To find a smart initialization, we first initialize the matrix $w^{(n)}$ in the ReLU-network parameter space and then use the forward mapping from ReLU-network to TTFS-network \cite{stanojevic2022} to set the weight matrix $W^{(n)}$.  
We call this the 'smart $\alpha$1 initialization'.
The other solution uses the $B1$-model which is the only model where weights in the TTFS-network and the ReLU-network are identical, see Eq. \eqref{eq:general_mapping} and Methods. We also refer to this model as \textit{identity mapping}. With both solutions the eigenvalues of the SNN Jacobian stay,  for the standard deep learning initialization, tightly within the unit circle, showing numerically that the vanishing-gradient problem is avoided at initialization (Fig.~\ref{fig:Fig2}b). 


\begin{figure*}[!t]
\centering
\includegraphics{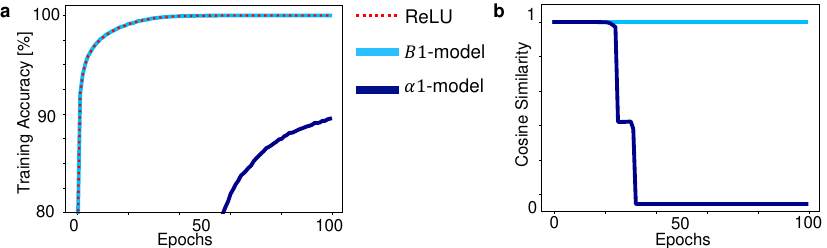}
\caption{{\bf Learning in 8-layer SNNs on the MNIST dataset for different models and initializations.} {\bf a.} A ReLU-network with standard deep learning initialization $w^{(n)}$ is trained (red-dashed). The $B1$-model (identity mapping $w^{(n)}=W^{(n)}$) with standard deep learning initialization follows the same training curve (light blue). The $\alpha1$-model ($w^{(n)}=\frac{W^{(n)}}{B_i^{(n)}}$) with 'smart $\alpha1$ initialization' deviates from the ReLU-network training curve (dark blue). {\bf b.} Cosine similarity measured during training between the weights of the ReLU-network and the weights of the (i) $B1$-model with standard deep learning initialization (ii) $\alpha1$-model with 'smart $\alpha1$ initialization'.} 
\label{fig:Fig3}
\end{figure*}

\subsection{The identity mapping makes training equivalent}

\label{sec:biased_traj}
Both the $\alpha1$-model with smart initialization and the $B1$-model described in the previous section avoid exploding gradients at initialization, but there is no guarantee that the same will hold during SNN training. To describe the gradient descent trajectory of the SNN, we consider a gradient descent step with learning rate $\eta$ when applying backpropagation to the TTFS-network: $\Delta W_{ij}^{(n)} = - \eta \dv{\mathcal{L}}{W_{ij}^{(n)}}$, and compute the corresponding update $\delta w_{ij}^{(n)}$ in the space of the ReLU-network parameters. 
Then $\delta w_{ij}^{(n)}$ can be expressed as $\delta w_{ij}^{(n)} = \mathcal{M}(W_{ij}^{(n)} - \eta \frac{d \mathcal{L}}{d W_{ij}^{(n)}}) - \mathcal{M}(W_{ij}^{(n)})$, where $w_{ij}^{(n)} = \mathcal{M}(W_{ij}^{(n)})$, see Eq. \eqref{eq:general_mapping}. 

\begin{table}
\centering
\begin{tabular}{l c l} 
\hline
\multicolumn{2}{c}{Model} & Test Acc [\%]  \\ 
\hline\hline 
\multicolumn{3}{l}{MNIST} \\
\hline\hline 
ReLU & FC2 & 98.30 \\
$\mathbf{SNN [ours]}$ & $\mathbf{FC2}$ & $\mathbf{98.30}$ \\
SNN \cite{comsa} & FC2 & 97.96 \\
SNN \cite{zhang21} & FC2 & 98 \\
SNN \cite{stanojevic2020} & FC2 & 98.2 \\
\hline
ReLU & FC16 & 98.43 $\pm$ 0.07\\
$\mathbf{SNN [ours]} $ & $\mathbf{FC16}$ & $\mathbf{99.42 \pm 0.07}$ \\
ReLU & VGG16 & 99.57 $\pm$ 0.01\\
$\mathbf{SNN [ours]}  $ & $\mathbf{VGG16}$ & $\mathbf{99.57 \pm 0.00}$ \\
\hline\hline 
\multicolumn{3}{l}{fMNIST} \\
\hline\hline 
ReLU & FC2 & 90.14 \\
$\mathbf{SNN [ours]}$ & $\mathbf{FC2}$ & $\mathbf{90.14}$ \\
SNN \cite{zhang21} & FC2 & 88.1 \\
SNN \cite{stanojevic2020}& FC2 & 88.93 \\
\hline
ReLU & LeNet5 & 90.9 $\pm$ 0.27\\
$\mathbf{SNN [ours]}$ & $\mathbf{LeNet5}$ & $\mathbf{90.94 \pm 0.21}$ \\
SNN \cite{zhang21} & LeNet5 & 90.1 \\
\hline
\end{tabular}
\caption{Performance after training an SNN using the $B1$-model (ours, bold face), compared to trained ReLU-networks and  SNN baselines.} 
\label{tab:table_equivalent}
\end{table}

With the B1-model, $\mathcal{M}(W_{ij}^{(n)})$ is the identity function, hence 
$\delta w_{ij}^{(n)} = \eta \dv{\mathcal{L}}{W_{ij}^{(n)}} =\Delta W_{ij}$. Therefore,  for the B1-model the training trajectories of SNN and ReLU networks are equivalent.
By contrast, for the $\alpha1$-model from \cite{stanojevic2022}, 
$\mathcal{M}(W_{ij}^{(n)})= (W_{ij}^{(n)})/(1+\sum_k W_{ik}^{(n)})$ is a nonlinear function (see Eq. \eqref{eq:general_mapping}) which leads to a difference between the $\delta w_{ij}^{(n)}$ from the reverse mapping and a direct ReLU-network update obtained through gradient descent $\Delta w_{ij}^{(n)} = \eta \dv {\mathcal{L}} {w_{ij}^{(n)}}$. Such a difference cannot be corrected with a different learning rate $\eta$, because the learning rate would have to be different not just for each neuron, but also for each update step (see Methods).
Hence, the gradient descent trajectory in the $\alpha1$-model  is systematically different than that of the equivalent ReLU model, even if their initializations are equivalent. A 
 specifically designed 'metric' \cite{surace20} that counterbalances update steps could be a solution but is not part of standard gradient optimization in machine learning.

In Fig. \ref{fig:Fig3} we illustrate the learning trajectories of the $B1$-model with standard deep learning initialization and $\alpha$1-model with 'smart $\alpha1$ initialization', as reported on the training data. Both TTFS-networks are initialized to be equivalent to the same ReLU-network. Nevertheless, we observe that only the $B1$-model follows the ReLU-network whereas the $\alpha$1-model diverges away despite a small learning rate.

\par To test our SNN training more broadly, we perform simulations of the $B1$-model for different architectures and compare them with earlier training approaches of TTFS-networks on MNIST and Fashion-MNIST (fMNIST) datasets. First, we consider a shallow network with one fully-connected hidden layer (FC2) \cite{Goltz21,zhang21, stanojevic2020, mostafa,comsa}. Before training, the ReLU-network and TTFS-network are initialized with the same parameters and the seed is fixed in order to avoid any other source of randomness. We compare our  
network (with 340 neurons in the hidden layer) with published networks  containing 340 or more neurons in the hidden layer. The test accuracy of our model  is higher than that of  all other SNNs (see Table~\ref{tab:table_equivalent}). 

Moreover, we tested a 16-layer fully-connected SNN (16FC), a 5-layer ConvNet SNN (LeNet5), and a 16-layer ConvNet SNN (VGG16). For deeper networks we noticed that, even though the SNN and ReLU network are initialized with the same parameters, sometimes they exhibit different performance after several epochs due to numerical instabilities. For this reason, in all cases where the number of hidden layers is larger than one we report the average performance across 16 learning trials with different random initial conditions. As expected from the theory, our SNN with identity mapping ($B1$-model) achieves the same performance as the ReLU-network (Table \ref{tab:table_equivalent}) and surpasses the test accuracy of previous works. Given these  results, all  experiments in the following sections are executed for SNNs with the identity mapping.

\subsection {Sparsity on large benchmarks}

\begin{figure}
  \begin{center}
    \includegraphics[width=0.3\textwidth]{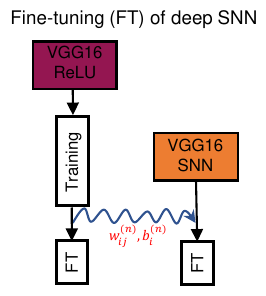}
  \end{center}
  \caption{{\bf Conversion followed by fine-tuning for a VGG16 network.}  A VGG16 SNN network (right)  is initialized (curly blue arrow)  with the weights and biases of a pretrained VGG16 ReLU network (left), and then fine-tuned (FT) with gradient descent using the framework of the $B1$-model. The pretrained ReLU network also undergoes further fine-tuning using standard BackProp optimization.  }
  \label{fig:Fig4}
\end{figure}

For a long time, tackling larger-scale image datasets like CIFAR100 \cite{cifar_dataset} or PLACES365 \cite{places365} (image size $224 \times 224 \times 3$, similar to ImageNet, but avoiding privacy concerns \cite{yang22q}) with TTFS-networks was considered impossible. To facilitate the training for these larger-scale datasets, we are going to combine conversion from pre-trained VGG16 ReLU models (step 1) and fine-tuning of the obtained SNN with gradient descent for the identity mapping (step 2, Fig. \ref{fig:Fig4}). A pre-trained ReLU model is  downloaded from an online repository \cite{cifar10_100, places365_model} and mapped to the SNN without any loss of performance (similarly to \cite{stanojevic2022}). Both networks are then fine-tuned for 10 epochs and 16 trials. 
%
The results that we obtain here through approximation-free learning are the
first one to close the performance gap in accuracy between deep TTFS-networks and deep ReLU-networks, see Table \ref{tab:table_large_sparsity}. 

The average fraction of spikes per neuron per data point ('SNN Sparsity') directly impacts the energy which is required for the SNN inference \cite{han2015deep,sacco2017,Davies21}. For example, in digital implementations, memory reads for accessing weights are expensive \cite{han2015deep}, but in a spiking implementation there is no need to access the weights of synapses that have not observed any spike. Similarly, in analog implementations spike transmission costs are likely to dominate energy consumption since 
the spike transmission cost $T_r$ increases significantly as the network scales to a larger number of neurons \cite{sacco2017}. Therefore, we expect the energy consumption to be dominated by  $T_r$
(see Methods). We thus aim to restrict the fraction of spikes per neuron by training with L1 regularization.

\begin{table*}[t!]
\centering
\begin{tabular}{l  c l l l l c} 
\\ \hline
  
 Dataset & Classes & \multicolumn{2}{c}{Test Acc [\%] w/o FT} & \multicolumn{2}{c}{Test Acc [\%] w/ FT} & SNN \\ 
  &  &  ReLU & SNN  & ReLU & SNN & Sparsity  \\ 
\hline\hline 
{\bf CIFAR10} &10& 93.59 \cite{cifar10_100} &93.59& {\bf 93.68\footnotesize{$\pm$0.02}} & {\bf 93.69\footnotesize{$\pm$0.00}} &{\bf 0.38}\\
CIFAR10 \cite{park2021training}&10&-&-&-&91.90&0.24\\
CIFAR10 \cite{zhou2021temporal}&10&-&-&-&92.68&0.62\\
\textbf{CIFAR10+L1}&{10}&92.82&92.82&\textbf{\textbf{93.28\footnotesize{$\pm$0.02}}} &\textbf{93.27\footnotesize{$\pm$0.02}} &\textbf{0.20}\\
{\bf CIFAR100} &100& 70.48 \cite{cifar10_100} & 70.48 & {\bf 72.23\footnotesize{$\pm$0.06}}& {\bf 72.24\footnotesize{$\pm$0.06}}&{\bf 0.38}\\
CIFAR100 \cite{park2021training}&100&-&-&-&65.98&0.28\\
\textbf{CIFAR100+L1}&{100}& 69.33 & 69.33 &\textbf{72.20\footnotesize{$\pm$0.04}}&\textbf{72.20\footnotesize{$\pm$0.04}}&\textbf{0.24}\\
\textbf{PLACES365}&{365}&52.69 \cite{places365_model}&52.69&\textbf{53.86\footnotesize{$\pm$0.01}}&\textbf{53.86\footnotesize{$\pm$0.02}}&\bf{0.54}\\
\textbf{PLACES365+L1}&365&48.67&48.67&\textbf{48.88\footnotesize{$\pm$0.06}}&{\bf 48.85}\textbf{\footnotesize{$\pm$0.06}}&{\bf 0.27}\\
\hline
\end{tabular}
\caption{Test accuracy and sparsity for a VGG16 architecture. The first column identifies the dataset and indicates whether L1 regularization was used. The second column gives the number of classes for each task; the third and fourth column show accuracy after pretraining (ReLU) and conversion (SNN) without fine-tuning (w/o FT); and the fifth and sixth column show the final results after fine-tuning (w/ FT). The right-most column presents the average number of spikes per neuron (sparsity). Final results obtained in this work are in bold.} 
\label{tab:table_large_sparsity}
\end{table*}

We first pretrain a ReLU network with L1-regularization and then transfer the weights to the SNN as initial condition for further fine-tuning over 16 trials, 10 epochs each. To allow a fair comparison, the ReLU network also undergoes the same fine-tuning procedure.  Table \ref{tab:table_large_sparsity} shows that  L1-regularization pushes the SNN Sparsity (mean across all trials) below 0.3 spikes per neuron. We conclude that the presented approach offers high-performance SNNs with very sparse spiking (as low as 0.2 spikes per neuron for CIFAR10) and therefore high energy-efficiency. Hence, it lends itself to a hardware implementation, where it can potentially serve as a low-power alternative to the state-of-the-art ANN solutions.

\subsection{Fine-tuning for hardware}
\label{sec:sim_mitigate}

In hardware, the high performance and sparsity of the SNN obtained in the software simulations may be affected by physical imperfections or constraints. Importantly,  our training algorithm can be used to fine-tune the SNN parameters given the specific hardware properties such as noise, quantization or latency constraints (see Methods for detailed explanation).
 
\begin{figure*}[!t]
\centering
\includegraphics{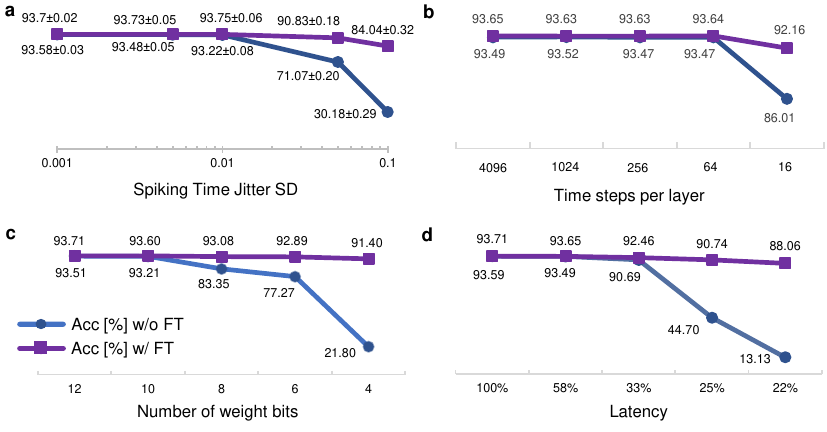}
\caption{{\bf Fine-tuning for noise robustness, quantization and latency.} VGG16 SNN on CIFAR10. In all cases after only 10 epochs of fine-tuning (w/ FT -- purple) the accuracy of the initially mapped network (w/o FT -- blue) is significantly improved. {\bf a.} Accuracy as a function of the  standard deviation (SD) of noise added to all spike times in the network (spiking time jitter). {\bf b.} Quantizing spiking times in the network to a given number of time steps per layer.  {\bf c.} Representing all weights $W_{ij}^{(n)}$ with given number of bits.  {\bf d.} Reducing the latency by reducing the ranges $[t_\text{min}^{(n)}, t_\text{max}^{(n)})$.}  
\label{fig:Fig5}
\end{figure*}

Let us consider  a ReLU-network that was pre-trained with full-precision weights, mapped to the SNN  and then deployed on an SNN device with noise, limited temporal resolution or limited weight precision. We test the success of fine-tuning in the presence of these constraints. To do so, we use 
a VGG16 architecture  (Fig.~\ref{fig:Fig5}a-c)
fine-tuned for 10 epochs on the CIFAR10 dataset. In all three cases (spike time jitter, time-step quantization or SNN weight quantization) fine-tuning enables to largely recover the performance of the unconstrained network. In particular, TTFS VGG16 networks achieve higher than $90\%$ test accuracy on CIFAR10 with 16 time-steps per layer or weights quantized to $4$ bits.

We also investigated whether it is possible to improve the classification latency through fine-tuning by reducing the intervals $[t_{\mathrm{min}}^{(n)},t_{\mathrm{max}}^{(n)})$ after conversion from ReLU-networks. Doing this naively, without fine-tuning, improves the latency, but the SNN performance drops well below that of the pre-trained ReLU-network. After fine-tuning, a test accuracy higher than $90\%$ is recovered, while  the latency can be improved up to a factor of $4$ (Fig.~\ref{fig:Fig5}d).

\section {Discussion}
The presented work provides a method to obtain high-performance sparse SNNs with the exact same performance as ANNs. We achieve this by identifying and  solving the challenging problem of vanishing-or-exploding gradients in spiking neural networks. The proposed identity mapping, which ensures that all spiking neurons reach the threshold with a trajectory of fixed slope, is crucial to ensure stability during learning by gradient descent. Furthermore, we have shown that with the identity mapping  training trajectories of ReLU-network and TTFS-network are equivalent. Our results have demonstrated that: training deep TTFS-networks yields identical performance to deep ReLU-networks on MNIST, Fashion-MNIST, CIFAR10, CIFAR100 and PLACES365 datasets; achieves high-sparsity crucial for energy efficiency; and enables fine-tuning to hardware constraints. Our work completely removes the performance disparity between deep SNNs and deep ANNs, outperforming all prior state-of-the-art research in 
training networks with TTFS coding.

The coding that we used is biologically inspired at a high level of abstraction, capturing specific principles from a plethora of neuroscientific findings.
In the brain, transient spiking activity initiated by short visual stimuli travels in a wave-like fashion along the visual processing pathway, with significant delays between visual areas \cite{Thorpe96,comsa, Richmond90,klimesch2012alpha, Yamins16}.  
In contrast to biology, where a neuron emits multiple spikes, our SNN model focuses only on the first spike of the transient response. Similar to biological spiking activity, we use a
continuous-time representation where a spike can occur at any moment when the membrane potential reaches the threshold \cite{mostafa, NEST, stanojevic2022,Gerstner02}. Our simulations are conducted with machine learning precision, similar to the ReLU network, setting the presented approach apart from discrete-time SNNs where spikes occur at the first time step after the threshold is reached. 

The high sparsity and high performance of the obtained SNN networks make them suitable for low-power hardware implementations. 
It is known that TTFS-networks can leverage the speed and energy-efficient characteristics of 
accelerators operating in analog \cite{Goltz21} as well as digital domain \cite{sandro2023}.
With our  method of training and fine-tuning SNNs, a potentially long classification latency or  sensitivity of model parameters to noise, which could negatively impact the metrics on device, are effectively mitigated. 
Finally, even though the obtained SNN models enforce a certain level of synchronicity, we do not believe that an implementation of TTFS-network like ours requires strict synchronization. What is important for our theory is that each layer roughly waits for the end of computations in the previous layer, but apart from that units and layers can function asynchronously. Even if every neuron experiences a small jitter in spike timing, it is possible to mitigate the effects  using our training algorithm.

In the future, our theory could serve as a starting point to address the training instabilities of fully-asynchronous SNNs  and to devise  hardware-friendly learning rules.
Remaining challenges that need to be addressed for a wider field of applications
are generalizations of our approach to skip-connections in ResNets and adaptations of TTFS networks to process a stream of temporally correlated input images, like a video.
Overall, we envision a process where pre-trained state-of-the-art ReLU-networks are converted into TTFS-networks and deployed on devices where they undergo continual online learning on the chip, ensuring energy-efficient and low-latency inference. 

\section*{Methods}

\paragraph{SNN neuron dynamics} Without loss of generality, we assume the potential $V_i^{(n)}$ to be unit-free and so are the parameters $W_{ij}^{(n)}$ and $
A_i^{(n)}, B_i^{(n)}$, whereas $t$ and $\tau_c$ have units of time. Rescaling time by $t\to(t/\tau_c)$ would remove the units, but we keep it in the equations to show the role of the conversion factor $\tau_c$. In biology $\tau_c$ in sensory areas is in the range of a few milliseconds \cite{temporal_retina,temporal_tactile}, whereas in hardware devices it could be in the range of microseconds or even shorter. 

\paragraph{Output layer} The output layer $N+1$ contains non-spiking read-out neurons. Each neuron $m$ simply integrates input spikes coming from layer $N$ during the time interval $[t_\text{min}^{(N)}, t_\text{max}^{(N)})$ without firing. Moreover, in this case, 
$A_i^{(N+1)}  = B_i^{(N+1)} - \sum_j W_{ij}^{(N+1)}$
 instead of 0. Integration of $V_m^{(N+1)}$ stops at time $t_\text{max}^{(N)}$, and the \textit{softmax} and the standard cross-entropy loss $\mathcal{L}$ are calculated using real-valued potentials, analogous to the real-valued activations of ANNs.

\paragraph{Proof of the exact reverse mapping from TTFS-network to ReLU-network}
\begin{proof}
Starting from the SNN definition in Eq. \eqref{eq:potential}, we compute analytically the spiking time $t_i^{(n)}$ of neuron $i$ in layer $n$. 
We consider the case $A_i=0$ for simplicity. We assume that the potential $V_i^{(n)}$ reaches the threshold $\vartheta_i^{(n)}$ at time  $t_i^{(n)}$  in the time window $[t_{\rm min}^{(n)},t_{\rm max}^{(n)})$. The spiking condition $\vartheta_i^{(n)}=V_i^{(n)}(t_i^{(n)})$ yields: 
\begin{equation}\label{eq:potential_integr}
\tau_c \vartheta_i^{(n)} =\sum_{j'}W_{ij'}^{(n)}\,(t_{\rm min}^{(n)}-t_{j'}^{(n-1)}) + B_i^{(n)}\,
(t_i^{(n)}-t_{\text{min}}^{(n)}).
\end{equation}
where $j'$ iterates over all neurons in layer $n-1$ which have generated a spike. 
For the spiking time $t_i^{(n)}$, we have:
\begin{equation}\label{eq:spiking_time}
B_i^{(n)} t_i^{(n)}= B_i^{(n)} t_{\rm min}^{(n)} +
\tau_c \vartheta_i^{(n)} - \sum_{j'}W_{i{j'}}^{(n)}(t_{\rm min}^{(n)}-t_{j'}^{(n-1)}) \, .
\end{equation}
We divide by the factor $B_i^{(n)}$ and subtract $t_{\text{max}}^{(n)}$ on both sides of Eq. \eqref{eq:spiking_time}, which yields
\begin{equation}
\label{eq:t_i}
t_i^{(n)} - t_{\text{max}}^{(n)}
=
\tau_c \frac{\vartheta_i^{(n)}}{B_i^{(n)}} + t_{\text{min}}^{(n)} - t_{\text{max}}^{(n)} +\sum_{j'} \frac{W_{i{j'}}^{(n)}}{B_i^{(n)}} (t_{j'}^{(n-1)} - t_{\text{min}}^{(n)}).
\end{equation}
Using  Eq. \eqref{eq:t_i}, one can now prove by induction that Eq. \eqref{eq:general_mapping} defines an equivalent ReLU network satisfying the identity 
$x_i^{(n)}  =   ({t}_{\rm max}^{(n)} - {t}_i^{(n)})/\tau_c $
for neurons that fire a spike. We set $x_i^{(n)}=0$ for neurons that do not fire a spike in the SNN
and note that the rectified linear unit $i$ is in its operating regime $x^{(n)}_i > 0$ if and only if the corresponding spiking neuron $i$ fires before  $t_{\rm max}^{(n)}$.
\end{proof}

\paragraph{Adaptive $t_{\rm max}^{(n)}$ parameters}
 $\vartheta_i^{(n)},t_{\rm min}^{(n)},t_{\rm max}^{(n)}$ are initialized such that the neurons which are active in the equivalent ReLU-network generate a spike in the interval $[t_{\rm min}^{(n)},t_{\rm max}^{(n)})$. Details of the choice of the base threshold and  $t_{\rm max}^{(n)}$ are given in \ref{ap:ttfs_details}. During training, as we update the network parameters $W_{ij}^{(n)}$ and $D_i^{(n)}$, the hyperparameters like $t_{\rm max}^{(n)}$ need to be changed, so that the condition that the neurons which are active in the equivalent ReLU-network generate a spike in the interval $[t_{\rm min}^{(n)},t_{\rm max}^{(n)})$ remains true. We suggest a new adaptive update rule which recalculates $t_{\rm max}^{(n)}$ such that all the spikes $t_i^{(n)}$ are moved away from the $t_{\rm max}^{(n)}$ boundary. Formally, when processing the training dataset, we update $t_{\rm max}^{(n)}$ as follows:
\begin{equation}
\label{eq:dervis_tmax}
\Delta t_\text{max}^{(n)} = 
 \begin{cases}
      \gamma (t_\text{max}^{(n)}-\text{min}_{i, \mu} t_i^{(n)}) - (t_\text{max}^{(n)} - t_\text{min}^{(n)}), & \text{if}\ t_\text{max}^{(n)}-t_\text{min}^{(n)} < \gamma (t_\text{max}^{(n)}-\text{min}_{i,\mu} t_i^{(n)})\\
      0, & \text{otherwise}
\end{cases}
\end{equation}
The minimum operator iterates over all neurons $i$ and input samples $\mu$ in the batch and $\gamma$ is a constant. After this update, we change the subsequent time window accordingly so that: $t_{\rm min}^{(n+1)} = t_{\rm max}^{(n)}$, and we iterate over all layers sequentially. The base threshold $\tilde{\vartheta}_i^{(n)}$ is then updated accordingly, see \ref{ap:ttfs_details}. For simplicity, in the theory section, we consider that this update has reached an equilibrium, so we consider that $t_{\rm max}^{(n)}$, $t_{\rm min}^{(n)}$, and $\tilde{\vartheta}_i^{(n)}$ are constants w.r.t. the SNN parameters, the condition $t_{\rm min}^{(n+1)} = t_{\rm max}^{(n)}$ is always satisfied and all the spikes of layer $n$ arrive within $[t_{\rm min}^{(n)},t_{\rm max}^{(n)})$. 

\paragraph{Calculating the SNN Jacobian}
In order to obtain the values of the $\dv{\mathbf{t}^{(n)}}{\mathbf{t}^{(n-1)}}$ matrix, we take the derivative of Eq. \eqref{eq:t_i}, which yields $\dv{\mathbf{t}^{(n)}}{\mathbf{t}^{(n-1)}} = M^{(n-1)} \cdot \frac{1}{B^{(n)}} \cdot W^{(n)}$ where $M^{(n-1)}$ is a diagonal matrix 
 with elements $M_{ij}^{(n-1)}=\delta_{ij} H(t_{\rm max}^{(n-1)}-t_i^{(n-1)})$  containing one  for all neurons $j'$ which generate a spike.

\paragraph{Exact reverse identity mapping}
The condition $B_i^{(n)}=1$ for all $i$ and $n$ (Fig. \ref{fig:Fig1}c), results in the identity mapping formula (see Eq. \ref{eq:general_mapping}):
\begin{equation}
\label{eq:linear_mapping}
w_{ij}^{(n)} \myeq W_{ij}^{(n)} 
\text{\hspace{1em} and \hspace{1em}}
b_i^{(n)} \myeq - \vartheta_i^{(n)} + \frac{t_{\text{max}}^{(n)} - t_{\text{min}}^{(n)}}{\tau_c}.
\end{equation}

\paragraph{SNN and ReLU training trajectories ($\delta w_{ij}^{(n)}$)} 
We calculate the update in ReLU-network parameter space $\delta w_{ij}^{(n)}$ as the difference between ReLU-network weights obtained from (i) the reverse-mapped updated SNN, i.e., $\mathcal{M}(W{_{ij}}^{(n)} - \eta \frac{d \mathcal{L}}{d W_{ij}^{(n)}})$ and (ii) the reverse-mapped original SNN, i.e., $\mathcal{M}(W{_{ij}}^{(n})$, where $w_{ij}^{(n)} = \mathcal{M}(W_{ij}^{(n)})$. 

For small learning rate $\eta$, we can employ a first-order approximation of the mapping function $\mathcal{M}$ around $W_{ij}^{(n)}$: $\mathcal{M}(W{_{ij}}^{(n)} - \eta \frac{d \mathcal{L}}{d W_{ij}^{(n)}}) \approx \mathcal{M}(W{_{ij}}^{(n}) - \eta  \dv{\mathcal{M}}{W_{ij}^{(n)}}\dv{\mathcal{L}}{W_{ij}^{(n)}}$, from which follows:

\begin{equation}
\label{eq:biased_trajectory}
\delta w_{ij}^{(n)} \approx - \eta  \dv{\mathcal{M}}{W_{ij}^{(n)}}\dv{\mathcal{L}}{W_{ij}^{(n)}}
= - \eta  
\left[\dv{w_{ij}^{(n)}}{W_{ij}^{(n)}}\right] ^2\, \dv{\mathcal{L}}{w_{ij}^{(n)}}, 
\end{equation}
where the second equality comes from plugging in $\dv{\mathcal{M}}{W_{ij}^{(n)}} = \dv{w_{ij}^{(n)}}{W_{ij}^{(n)}}$ and $\dv{\mathcal{L}}{W_{ij}^{(n)}} = \dv{\mathcal{L}}{w_{ij}^{(n)}}\dv{w_{ij}^{(n)}}{W_{ij}^{(n)}}$. For the $\alpha1$-model, the difference between the $\delta w_{ij}^{(n)}$ and a direct ReLU-network update $\Delta w_{ij}^{(n)} = \eta \dv {\mathcal{L}} {w_{ij}^{(n)}}$ cannot be corrected with a different learning rate $\eta$. This is due to the fact that the multiplicative bias which appears in Eq. \eqref{eq:biased_trajectory} changes for every neuron pair $(i,j)$ and algorithmic iteration, i.e. $\dv{w_{ij}^{(n)}}{W_{ij}^{(n)}} = \dv{\mathcal{M}_{\boldsymbol{\alpha}}}{W_{ij}^{(n)}} = \frac{B_i^{(n)} - W_{ij}^{(n)}}{(B_i^{(n)})^2}$. 

\paragraph{Simulation details}
Each simulation run was executed on one NVIDIA A100 GPU. In all experiments $\tau_c$ was set to $1 \mathcal{U}$ where $\mathcal{U}$ stands for the concrete unit such as $ms$ or $\mu s$. Note that although the choice of units in simulations can be arbitrary, it becomes a critical parameter for a hardware implementation. Moreover, we set $\zeta=0.5$ whereas a hyperparameter $\gamma=10$ ensures that even for higher values of initial learning rate the neurons, which are active in the equivalent ReLU-network, generate a spike in the interval $[t_{\rm min}^{(n)},t_{\rm max}^{(n)})$. The simulation results were averaged across 16 trials. Batch size was set to 8. In all cases we used the Adam optimizer where for the initial learning rate ``$\mathrm {lr}_0$'' and iteration ``$\mathrm {it}$'' an exponential learning schedule was adopted following the formula: $\mathrm {lr}_0 * 0.9^\frac{\mathrm{it}}{5000}$.

\par The data preprocessing included normalizing pixel values to the [0, 1] range and, in the case of a fully-connected network, the input was also reshaped to a single dimension. For MNIST, Fashion-MNIST, CIFAR10 and CIFAR100 the training was performed on the training data, whereas the evaluation was performed on the test data. For PLACES365 the fine-tuning was performed on $1\%$ random sample of the training data and the evaluation was performed on the validation data (since the labels for test data are not publicly available). 

\par In fully-connected architectures all hidden layers contain 340 neurons. The LeNet5 contains three convolutional, two max pooling and two fully-connected layers with 84 and 10 neurons, respectively. Moreover, some of the datasets utilize a slightly modified version of VGG16. The kernel was always of size 3 and the input of each convolutional operation was zero padded to ensure the same shape at the output. For MNIST dataset, due to a small image size, the first max pooling layer in VGG16 was omitted. In this case there are two fully-connected hidden layers containing 512 neurons each. For CIFAR10 and CIFAR100 the convolutional layers are followed by only one fully-connected hidden layer containing 512 neurons, yielding 15 layers in total. Finally, for PLACES365, there are two fully-connected hidden layers with 4096 neurons each. The spiking implementation of max pooling operation was done as in \cite{stanojevic2022}.

\label{ap:simulation}

\paragraph {Simulations details for demonstrating TTFS-SNN and ReLU training trajectories}
In Fig.~\ref{fig:Fig3} we illustrated that training $\alpha1$-model with 'smart $\alpha1$ initialization' is difficult. For the optimization process we used plain stochastic gradient descent (SGD). In Fig. \ref{fig:Fig3}a the $B1$-model was trained with initial learning rate equal to 0.0005, which is the same as in the corresponding ReLU network. For $\alpha1$-model with 'smart $\alpha1$ initialization' the learning process with the same initial learning rate struggles to surpass the training accuracy of around $20\%$. In this case we found the optimal initial learning rate to be 0.00003, leading to a slower training compared to both ReLU-network and $B1$-model. In Fig. \ref{fig:Fig3}b the goal is to understand how much the SNN weights diverge from the ReLU weights during training. In order to enable a fair comparison in this case, all three networks were trained with initial learning rate equal to 0.00003. 

\paragraph {Simulations details for large benchmarks} 
Some of the pretrained models we used have batch normalization layers \cite{cifar10_100}. The exact mapping fuses them with neighbouring fully-connected and convolutional layer similar as in \cite{stanojevic2022}, after which the fine-tuning is conducted for 10 epochs. Since the models are already pretrained, the fine-tuning is done with a reduced initial learning rate of $10^{-6}$ for CIFAR10 and CIFAR100, and $10^{-7}$ for PLACES365. Importantly, the simulations show that the SNN fine-tuning yields zero performance loss compared to the corresponding ReLU network. 

\paragraph{Estimation of the dominant factor of energy consumption in analog hardware}
If we assume that the neurons are implemented with capacitors of capacitance $C$, which are being charged as the input spikes arrive, then the overall energy for processing a data point can be estimated as $\sum_{i', n} (T_r + 0.5(\vartheta_{i'}^{(n)})^2C) + 0.5 \sum _{i'',n}(V_{i''}^{(n)}(t_{\rm{max}}^{(n)}))^2C$. Here the transmission cost per spike is denoted by $T_r$, whereas the other two terms describe the charging cost of the capacitor. The index $i'$ runs over all neurons which fire a spike. For these neurons we add the transmission cost to the charging energy,  calculated simply using the threshold value $\vartheta_{i'}^{(n)}$. Analogously, for all neurons that do not spike (index $i''$) the charging energy  is calculated using the value of the potential $V_{i''}^{(n)}$ at time instant $t_{\mathrm{max}}^{(n)}$, which is smaller than the threshold $\vartheta_{i''}^{(n)}$. Therefore, when a neuron spikes it contributes a larger share to the energy consumption than when it stays silent. 
To estimate the average capacitor energy 
per neuron we use the definition $\theta^2 = (1/N) \sum_i (\vartheta_i^{(n)})^2$ where the sum runs over all neurons in all layers. To find out how much the charging cost is reduced if a neuron does not spike we calculate the relative fraction $r=(1/N)\sum r_i$ where $r_i = 
\langle (V_{i}^{(n)}(t_{\rm{max}}^{(n)}))^2 \rangle / (\vartheta_i^{(n)})^2$. On CIFAR 10 and CIFAR 100 we find values in the range $0.8<r<0.85$ which suggest that the relative reduction of capacity energy for non-spiking versus spiking is in the range of ten to twenty percent. 
In general, the spike transmission cost $T_r$ increases significantly as the network scales to a larger number of neurons \cite{sacco2017}, therefore, we expect the energy terms to be related as $T_r \gg 0.5\theta^2C > 0.5r\theta^2C$. Hence we expect sparsity to significantly reduce transmission cost and only marginally reduce charging cost.

 \paragraph {Simulation details for fine-tuning for hardware}
We implement independently four types of hardware constraints: (i) spiking time jitter, (ii) reduced number of time steps per layer, (iii) reduced number of weight bits, and (iv) latency limitations. In practice, these constraints often coexist, but this is not considered here. 

 \par \textit{(i) Spiking time jitter} (Fig. \ref{fig:Fig5}a) A Gaussian noise of given standard deviation is added to the spiking times in each layer. 

 \par  \textit{ (ii) Time quantization} (Fig. \ref{fig:Fig5}b) In digital hardware, the spike times of the network are subjected to quantization leading to discrete time steps. To mitigate the impact of the spike time outliers, the size of the $[t_\text{min}^{(n)}, t_\text{max}^{(n)})$ interval is chosen to contain $99\%$ of the activation function outputs in layer $n$ when training data is sent to the input of the ReLU network. 
 The result is that some neurons in layer $n$ can fire a spike too early, i.e., before $t_\text{min}^{(n)}$. In our software implementation, the input of  'early' spiking times is treated as if the spike had occurred  at $t_{\rm min}^{(n)}$.  The initial interval $[t_\text{min}^{(n)}, t_\text{max}^{(n)})$  is divided into quantized steps, which are fixed during the fine-tuning. In other words,  in this case the adaptive rule which changes $t_{\rm max}^{(n)}$ is not applied. 
 
 \par \textit{(iii) Weight quantization} (Fig. \ref{fig:Fig5}c) To reduce the size of the storage memory, we apply quantization-aware training such that at the inference time the weights are represented with a smaller number of bits. Similarly as for the spiking time, we remove outliers before the quantization. In this case, we remove a predefined percentile as follows on both sides of the distribution. In case of a larger number of bits, only the first and last percentile were removed. However, in case of a 4-bit representation, we reduce the interval further by removing first four and last four percentiles. As before, the obtained range is divided into quantized steps, which are then fixed during the fine-tuning. At the inference time the quantized steps are scaled to the integer values on $[-2^{q-1}, 2^{q-1}-1]$ range (where $q$ is the number of bits), whereas the other parameters are adjusted accordingly. 

 \par \textit{(iv) Reduced latency} (Fig. \ref{fig:Fig5}d) The robustness to a reduced classification latency is tested by picking smaller $[t_\text{min}^{(n)}, t_\text{max}^{(n)})$ intervals. We emphasize that the adaptive rule which changes $t_{\rm max}^{(n)}$ is not applied here, i.e. the interval is fixed during fine-tuning. Note that with a reduced interval, it may happen that a neuron in layer $n$ fires before  $t_\text{min}^{(n)}$. If so, the input that it causes in layer $n+1$ is, in our implementation, only taken into account for 
 $t>t_\text{min}^{(n)}$; cf. Eq. \eqref{eq:potential}.
 To balance the relative reduction of interval across  all layers, the reduced $[t_\text{min}^{(n)}, t_\text{max}^{(n)})$ interval is obtained such that it contains some percentage of the activation function outputs in layer $n$  of the equivalent ReLU network when training data is used as input. The chosen values of percentiles are 100, 99, 95, 92 and 90 (yielding 22\% of the initial latency). 
 For convolutional layers we keep $\zeta=0.5$, and otherwise we use $\zeta=0$. 
 

\vspace{1cm}
\textbf{Acknowledgements}
The research of W.G. and G.B. was supported by a Sinergia grant (No CRSII5 198612) of the Swiss National Science Foundation. Fig. \ref{fig:Fig1}a has been created with BioRender.com. We would like to thank our colleagues from the IBM Emerging Computing \& Circuits team for discussions. \\

\textbf{Author contributions} A. S., W. G., S. W., G. B., G. C. and A. P. contributed conceptually. W. G. conceived
the idea. A. S., G. B. and S. W. developed the theory. A. S. designed and performed the simulations. A. S. and G. B. wrote the manuscript with input from W. G., S. W., G. C. and A. P.\\

{\bf Competing interest}.
The authors declare no competing interests. \\

\textbf{Data availability}
The datasets utilized during the current study are available in
public repositories. 

\bibliographystyle{naturemag}
\bibliography{library-CLEAN}

\newpage


\begin{center}
{\LARGE \bf Supplementary Information}
\end{center}
\newcounter{allsections}
\setcounter{allsections}{\value{section}}
\renewcommand{\thesection}{Supplementary Note \the\numexpr\value{section}-\value{allsections}\relax}
\titleformat{\section}[block]{\normalfont\large\bfseries}{\thesection: }{1em}{}

\newcounter{allfigures}
\setcounter{allfigures}{\value{figure}}
\renewcommand{\thefigure}{\the\numexpr\value{figure}-\value{allfigures}\relax}
\renewcommand{\figurename}{Supplementary Figure}

\section{Generalization to other neuronal dynamics}
\label{ap:generalization}

\begin{figure*}[!h]
\centering
\includegraphics{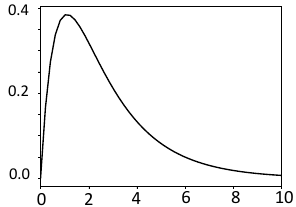}
\caption{An example of $a(t) = [1 - \mathrm{exp}(-\frac{t}{\tau_1})] \mathrm{exp}({-\frac{t}{\tau_2}})$ function for $\tau_1=1, \tau_2=2$.}
\label{fig:Fig6}
\end{figure*}

\paragraph{Linearization of the double exponential} In this paper we solve the vanishing-gradient problem for a spiking neural network with piecewise linear postsynaptic potential, i.e. an input spike at time $t_j^{(n-1)}<t$ causes the following response $a_i^{(n)}(t)$ in neuron $i$ of layer $n$:
\begin{equation}
\label{eq:a_lin}
a_i^{(n)}(t)=\frac{t-t_j^{(n-1)}}{\tau_c}H(t-t_j^{(n-1)}).
\end {equation}
However, biologically inspired models in related works \cite{comsa, Goltz21} often use a double-exponential filter, i.e.:
\begin{equation}
\label{eq:a_exp}
a_i^{(n)}(t) = [1 - \mathrm{exp}(-\frac{t-t_j^{(n-1)}}{\tau_1})] \mathrm{exp}(-\frac{t-t_j^{(n-1)}}{\tau_2}) H(t-t_j^{(n-1)}),
\end{equation}
where $\tau_1$ and $\tau_2$ are time constants and $\tau_2\geq 2\tau_1>0$ (Supplementary Fig. \ref{fig:Fig6}). 

We notice that in the vicinity of zero, the exponential function can be expressed using Taylor expansion as: $\mathrm{exp}({-\frac{t}{\tau}}) = [1 - \frac{t}{\tau}  + (1/2) (\frac{t}{\tau})^2 \dots ]$. Therefore the equation for $a_i^{(n)}(t)$ of the double-exponential filter can be approximated around 0 as:
\begin{equation}
\label{eq:a_exp_approx}
a_i^{(n)}(t)=\frac{t-t_j^{(n-1)}}{\tau_1}H(t-t_j^{(n-1)}).
\end{equation}

To show the relationship between this linearized model we first have to set constant $\tau_c=\tau_1$ in the neuron dynamics in Eq. \eqref{eq:potential} to match the physically interpretable time constant $\tau_1$. 
As a result the potential $V_i^{(n)}$ of neuron $i$ in layer $n$ at time $t$ becomes:
\addtocounter{equation}{1}
\begin{equation}\label{eq:pot_alpha_funct}
V_i^{(n)} (t)  = \left\{
\begin{split}
\; A_i^{(n)}& \frac{t - t_{\rm{min}}^{(n-1)}}{\tau_1} +
\sum_j W_{ij}^{(n)}  \frac{t - t_j^{(n-1)}}{\tau_1}  H(t-t_j^{(n-1)}) &  \quad \text{\rm for \quad}  t \le t_{\text{min}}^{(n)} \\ 
\; B_i^{(n)}& \frac{t - t_{\rm{min}}^{(n)}}{\tau_1}
+ V_i^{(n)} (t_{\rm{min}}^{(n)})
\nonumber
&  \quad \text{\rm for \quad} t_{\text{min}}^{(n)} < t \le t_{\text{max}}^{(n)} 
\end{split}
\qquad \right. ({\arabic{equation}})
\end{equation}
An analogy with the linearized model is therefore possible if the values $A_i^{n}$ and $W_{ji}^{(n)}$ represent the rising slopes of post-synaptic potentials for a more biological neuron. The remaining difference with the linearized model arises in the second phase, i.e. the interval $t_{\text{min}}^{(n)} < t \le t_{\text{max}}^{(n)}$, which does not follow a classical integrate and fire model. 
To reconcile our two-phase model with a more biologically plausible single-phase integrate and fire model, we can assume that all neurons of layer $(n)$ that did not spike before $t_{\text{min}}^{(n)}$ are forced to spikes at time $t_{\text{min}}^{(n)}$ (for instance with a strong excitatory input shared for the entire layer). In this way, continuing the first phase after $t_{\text{min}}^{(n)}$ yields the dynamics: $V_i^{(n)} (t) = (A_i^{(n)} + \sum_j W_{ji}) \frac{t - t_{\rm{min}}^{(n)}}{\tau_1}
+ V_i^{(n)} (t_{\rm{min}}^{(n)})$ which is equivalent to a our two-phase model as long as $B_i^{(n)}=A_i^{(n)} + \sum_j W_{ji}$.
This analogy enables the extrapolation of the theoretical conditions for stable gradient descent optimization to the linearized model: the condition $B_i^{(n)}=1$ yields the constraint $A_i^{(n)} + \sum_j W_{ji}=1$ in a plausible single phase model. Outside of the linearized setting, we conjecture from our theoretical analysis that the recipe to propagate gradients robustly is generally to cross the threshold with slope $1$.

\paragraph{Scaling the interval
$[t_\mathrm{min}^{(n)},t_\mathrm{max}^{(n)})$ to stay in the linear range}
\par In order for the network dynamics to always remain in the linear ramping phase of  the double-exponential filter, we need to ensure that the linear approximation given in Eq. \eqref{eq:a_exp_approx} is satisfied within the entire coding interval $\text{max}_n[t_\mathrm{max}^{(n)} - t_\mathrm{min}^{(n-1)})$. As the maximum of the function is reached at $t>\tau_1$ (Supplementary Fig. \ref{fig:Fig6}), we require for the implementation of our SNN with a double-exponential model that:
\begin{equation}
\label{eq:alpha_cond}
\text{max}_n[t_\mathrm{max}^{(n)} - t_\mathrm{min}^{(n-1)}]< 0.5 \tau_1.
\end{equation}

This is possible if we separate the definition of the interval $\tau_c$ of the pixel encoding from the neuron time constant $\tau_1$ (see \ref{ap:ttfs_details} for the recursive construction of the intervals $[t_\mathrm{max}^{(n)} - t_\mathrm{min}^{(n-1)})$). Keeping the notation $\tau_1$ for the neuron time constant and $\tau_c$ for the pixel encoding, we can construct our network with an arbitrary scaling factor between them to fulfill Eq.~\eqref{eq:alpha_cond}. 

\paragraph{Example for a CIFAR10 network}
\par In order to observe what could be concrete values satisfying the condition given in Eq. \eqref{eq:alpha_cond}, let's take as an example the CIFAR10 dataset and VGG16 architecture,  already explored in detail in the section on fine-turning for hardware. For the model that was fine-tuned for reduced latency (Fig. \ref{fig:Fig5}d), the $[t_\mathrm{min}^{(n-1)}, t_\mathrm{max}^{(n)})$ interval has a value of around $10\tau_c$. Therefore, the acceptable value for $\tau_1$ is $\tau_1=20\tau_c$. 

\par In addition to minor mismatches between the linear and double-exponential postsynaptic potential, caused by the above approximation steps, further mismatches may arise from the heterogeneities of the hardware and other hardware constraints. It 
is likely that fine-tuning the model, as explained in the main text, would become crucial in this setting.

\section{Setting $t_{\rm max}^{(n)}$  and the threshold} 
\label{ap:ttfs_details}

\paragraph{Initialization}
\par As indicated in  the main text, the base threshold $\Tilde{\vartheta}_i^{(n)}$
and the parameter $t_{\rm max}^{(n)}$ are initialized recursively starting from the input interval $[t_{\rm min}^{(0)}, t_{\rm max}^{(0)})$.
Let us now assume that we have adjusted the base threshold and the maximum firing time
$t_{\rm max}^{(n-1)}$
up to layer $n-1$.

In layer $n$, the earliest firing time  $t_{\rm min}^{(n)}$ is defined as:
$t_{\rm min}^{(n)}=t_{\rm max}^{(n-1)}$.
At time $t_{\rm min}^{(n)}$
we evaluate the membrane potential of all neurons in layer $n$ and determine its maximum:
$\max_{\mu, i} V_i^{(n)}(t_{\rm min}^{(n)}) $, where the index $\mu$ is running over many samples from the training dataset and $i$ iterates over all neurons in layer $n$.
Since, for all valid mappings, the slope factor $B_i^{(n)}$
of trajectories is positive for $t>t_{\rm min}^{(n)} $,
we define a reference potential in layer $n$ as:
\begin{equation}
  \label{eq:basethrehold}
   \Tilde{V}_0^{(n)} = (1+\zeta)  \max_{\mu, i} V_i^{(n)}(t_{\rm min}^{(n)}),
\end{equation}
where $\zeta>0$ is a small safety margin. 
We choose the latest possible firing time $t_{\rm max}^{(n)}$ to be
\begin{equation}
  t_{\rm max}^{(n)} \myeq t_{\rm min}^{(n)} + \tau_c \Tilde{V}_0^{(n)}/B_0,
\end{equation}
where $B_0$ is a reference slope factor of unit value.
We then set the base threshold for neuron $i$ in layer $n$ to
\begin{equation}
\label{eq:tilde_theta_2}
\Tilde{\vartheta}_i^{(n)}
\myeq B_i^{(n)} \left(\frac{ t_{\rm max}^{(n)}
  - t_{\rm min}^{(n)}}{\tau_c} \right).
\end {equation}
Since the neuron-specific
threshold $\vartheta_i^{(n)} \myeq \Tilde{\vartheta}_i^{(n)} + D_i^{(n)}$
is initialized with a shift parameter $D_i^{(n)}=0$,
the choice in Eqs. (\ref{eq:basethrehold}) -- (\ref{eq:tilde_theta_2}) guarantees that, with our initialization of the network parameters, the threshold is reached at a time $t>t_{\rm min}^{(n)}$ from below.

Note that Eq. (\ref{eq:tilde_theta_2}) defines the base threshold for $t>t_{\rm min}^{(n)}$. Since for $t<t_{\rm min}^{(n)} $
the membrane potential trajectories could transiently take a value above
$\Tilde{\vartheta}_i^{(n)}$,
we formally set the threshold for $t<t_{\rm min}^{(n)} $ 
to a large value (e.g., $100 \cdot \Tilde{\vartheta}_i^{(n)}$) so as to make spiking impossible \cite{stanojevic2022}.

For the identity mapping between TTFS-network and ReLU-network, which is the one chosen to avoid the vanishing-gradient problem, the actual slope factor takes a value $B_i^{(n)}=B_0=1$. In the context of this mapping,
 we note that $V_i^{(n)}(t)$ at time $t_\mathrm{min}^{(n)}$ has the same value as the activation variable of neuron $i$ in layer $n$ of the equivalent ReLU network, see Eqs. \eqref{eq:potential} and \eqref{eq:general_mapping}. Therefore the interval $[t_{\rm min}^{(n)}, t_{\rm max}^{(n)})$  is large enough to encode all outputs of layer $n$ in the ReLU-network at initialization (with bias parameter initialized at zero).

\paragraph{Iterative updates during training}
\par
Throughout training the $t_\mathrm{max}^{(n)}$ and the base threshold $\Tilde{\vartheta}_i^{(n)}$ are related by Eq. (\ref{eq:tilde_theta_2}).
In each iteration the parameters $W_{ij}^{(n)}$ and $D_i^{(n)}$ change. Whenever necessary, the iterative
update rule
for $t_{\rm max}^{(n)}$ 
in Eq. (\ref{eq:dervis_tmax})
shifts the maximal firing time in a regime with additional safety margin.
This influences in turn the
threshold $\vartheta_i^{(n)}$ which is recalculated according to Eq. (\ref{eq:tilde_theta_2}). 

\begin{figure*}[t]
\centering
\includegraphics{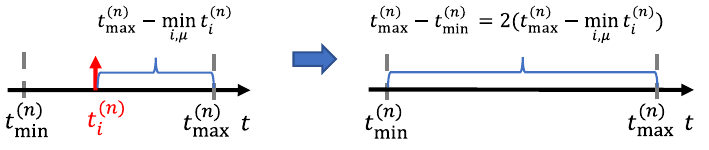}
\caption{{\bf Updating the latest spike time  ${t_\mathrm{max}^{(n)}}$ using $\gamma=2$}. If the minimal spiking time across neurons $i$ and batch inputs $\mu$ lies in the first half of the interval $[t_\text{min}^{(n)}, t_\text{max}^{(n)})$, the interval is extended.} 
\label{fig:Fig7}
\end{figure*}

\paragraph{Additional Remarks}
\par
(i) In principle, we are free to initialize $t_{\rm max}^{(n)}$ (or $\Tilde{\vartheta}_i^{(n)}$)
at arbitrarily high values, much larger than  those proposed above. In this case the adaptive rule for $t_{\rm max}^{(n)}$ (Supplementary Fig. \ref{fig:Fig7}) can be omitted. However, the trade-off becomes a very long spiking delay, in particular in networks with many layers.

\par Both the initialization of the reference potential with a parameter $\zeta>0$ and the iterative update rule
for $t_{\rm max}^{(n)}$ 
in Eq. (\ref{eq:dervis_tmax}) provide a safety margin that leads to spiking delays that could potentially be avoided. However, the $t_{\rm max}^{(n)}$ used during training does not need to be same as during inference.
In order to reduce the classification latency during inference, $t_{\rm max}^{(n)}$
is recalculated with the fixed parameters found after training such that, in each layer $n$, the earliest possible spike across all neurons and a representative sample from the training base happens immediately after $t_{\rm min}^{(n)}$ which in turn leads to a tight value for the threshold via Eq. (\ref{eq:tilde_theta_2}).

\end{document}